\documentclass[letterpaper, conference]{ieeeconf}

\IEEEoverridecommandlockouts                              %

\let\labelindent\relax
\usepackage{subcaption}
\usepackage{enumitem}
\usepackage{times}
\usepackage{epsfig}
\usepackage{graphicx}
\usepackage{amsmath}
\usepackage{amssymb}
\usepackage{booktabs}
\usepackage[table,xcdraw]{xcolor}
\usepackage{placeins}
\usepackage{float}
\usepackage{tabularx}
\usepackage{multirow}
\usepackage{siunitx}
\usepackage{scalerel}
\usepackage{balance}
\usepackage{url}
\usepackage{optidef}

\usepackage[left=0.75in,right=0.75in,bottom=0.75in,top=0.75in]{geometry}

\usepackage{bm}

\usepackage{cite}

\usepackage{color}
\definecolor{mygreen}{RGB}{26, 148, 49}

\usepackage{float}
\usepackage[multiple]{footmisc}

\begin{document}
\title{\LARGE \bf Robust Fusion for Bayesian Semantic Mapping}

\author{David Morilla-Cabello, Lorenzo Mur-Labadia, Ruben Martinez-Cantin, and Eduardo Montijano\\
\thanks{All authors are with the Instituto de Investigaci\'on en Ingenier\'ia de Arag\'on, Universidad de Zaragoza, Spain 
\texttt{\small \{davidmc, lmur, rmcantin, emonti\}@unizar.es}}
\thanks{This work has been supported by Spanish projects PID2021-125514NB-I00, PID2021-125209OB-I00, TED2021-129410B-I00 and TED2021-131150B-I00 funded by MCIN/AEI/10.13039/501100011033, by ERDF A way of making Europe and by the European Union NextGenerationEU/PRTR and Spanish grant FPU20-06563.}
}
\maketitle
\thispagestyle{empty}
\pagestyle{empty}
\begin{abstract}
The integration of semantic information in a map allows robots to understand better their environment and make high-level decisions. In the last few years, neural networks have shown enormous progress in their perception capabilities. However, when fusing multiple observations from a neural network in a semantic map, its inherent overconfidence with unknown data gives too much weight to the outliers and decreases the robustness. To mitigate this issue we propose a novel robust fusion method to combine multiple Bayesian semantic predictions. Our method uses the uncertainty estimation provided by a Bayesian neural network to calibrate the way in which the measurements are fused. This is done by regularizing the observations to mitigate the problem of overconfident outlier predictions and using the epistemic uncertainty to weigh their influence in the fusion, resulting in a different formulation of the probability distributions. We validate our robust fusion strategy by performing experiments on photo-realistic simulated environments and real scenes. In both cases, we use a network trained on different data to expose the model to varying data distributions. The results show that considering the model's uncertainty and regularizing the probability distribution of the observations distribution results in a better semantic segmentation performance and more robustness to outliers, compared with other methods. \\\textit{Video} - \url{https://youtu.be/5xVGm7z9c-0}
\end{abstract}

\section{Introduction}
Robots rely on understanding their surroundings to work autonomously and make informed decisions. Semantic information is critical for enabling reasoning on a higher abstraction level in complex tasks such as recognizing different objects~\cite{duan2023semantic}, driving safely~\cite{geiger2012cvpr} or helping in our homes~\cite{wang2020home}. Neural networks have played an important role in making this possible, but their application in robotic perception pipelines still presents multiple challenges~\cite{sunderhauf2018limits}. In this paper, we address the problem of robustness when combining multiple, potentially biased, neural network predictions in a semantic mapping pipeline. 

\begin{figure}[!t]
    \centering
    \includegraphics[width=0.9\columnwidth]{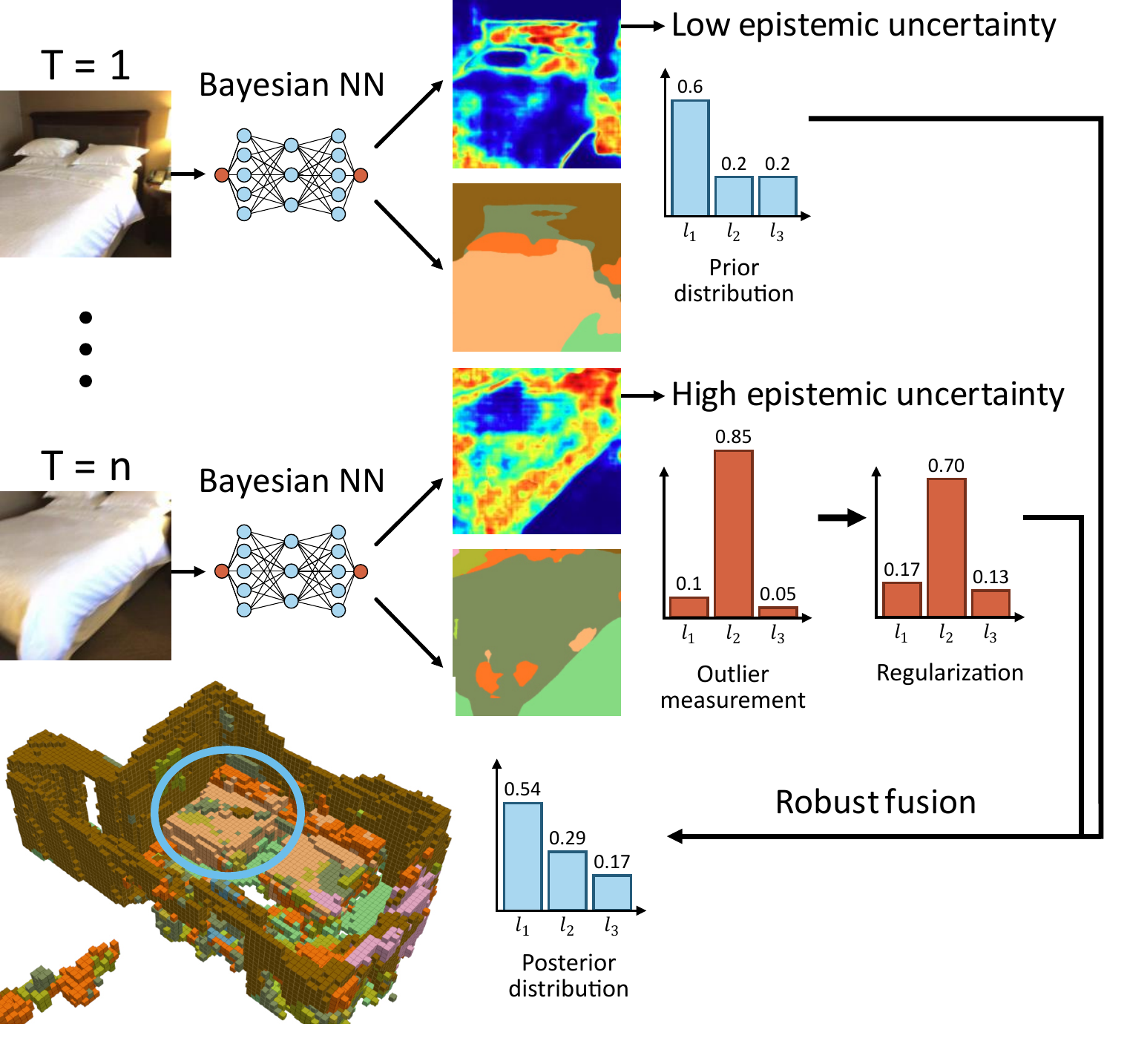}
    \caption{\footnotesize{We introduce a robust fusion method to perform semantic mapping. Since neural networks are overconfident sensors, they can output misclassified predictions with high confidence. Using Bayesian neural networks, we regularize its output and weigh observations according to their epistemic uncertainty, obtaining a fusion method robust to outlier detections.}}
    \label{fig:my_label}
\end{figure}

Compared to traditional sensors, which are well understood and calibrated, the behavior of neural network models still lacks interpretability. 
Firstly, neural network predictions assume that they are always within the trained distribution and are agnostic to the real distribution of the input data. Secondly, even if the output of traditional neural networks can be understood as a probability (i.e., classification confidence among a set of classes), these are often overconfident and produce misclassified predictions with a high confidence level \cite{guo2017calibration}. 
Despite this, existing semantic fusion methods~\cite{McCormac2017semanticfusion,Sunderhauf2017meaningful,Rosinol2020kimera,Asgharivaskasi2021active} fully trust the neural network predictions, which makes them vulnerable to the aforementioned limitations.

To prevent these problems, the main contribution of this work is a new fusion method that exploits the information given by a Bayesian neural network to increase the robustness of the resulting semantic map. Recent advances in Bayesian deep learning offer tools that enable reasoning about the predictions of neural networks in the form of uncertainties or confidences. These uncertainties have been successfully utilized in embodied active learning~\cite{chaplot2021seal, nilsson2021embodied, ruckin2022informative} and domain adaptation \cite{zurbrugg2022embodied}, but their potential in calibrating neural networks as sensors and enhancing the robustness of semantic information acquisition at test time has scarcely been investigated. In that sense, our method uses the uncertainties provided by a Bayesian neural network to (i) apply a regularization in the observations to reduce the influence of prediction bias and (ii) represent the network prediction as a Dirichlet distribution to incorporate the lack of knowledge from the network, which allows us to fuse the data considering the right uncertainty of each measurement. We validate our method in simulated and real environments, showing improvements with respect to non-robust approaches.

\section{Related Work}

\subsubsection{Semantic Mapping}

There are several ways to use neural networks for semantic mapping, including end-to-end, averaging, and fully Bayesian methods. For the first type, the data association and update of the semantic probabilities are learned in an end-to-end pipeline including recurrent structures to account for the temporal dimension \cite{Xiang2017darnn,Yang2021tupper}. Even though these are interesting approaches, end-to-end pipelines are complex to analyze and require specific training.

Averaging methods aggregate directly the predictions obtained from the neural network. In some cases, this is done considering only the winning class in each observation and performing a voting scheme~\cite{grinvald2019volumetric,Schmid2022multitsdf}.
In other cases, the probabilities given by the network are averaged for the total number of observations~\cite{Sunderhauf2017meaningful,McCormac2018fusionplusplus}.
The former is motivated by the excessive confidence typically shown by semantic segmentation neural networks.
While these approaches are very efficient, Bayesian methods provide a better framework to handle uncertainties and probability distributions. 

Bayesian methods offer an interesting framework to combine the prediction of the network in a probabilistic fashion.
Among these types of works, Semantic Fusion \cite{McCormac2017semanticfusion} updates the probability distribution of the \emph{surfels} generated by their SLAM system using a recursive Bayesian update. This is also used by current semantic mapping approaches \cite{Rosinol2020kimera}. Similarly, the work by Asgharivaskasi et al. \cite{Asgharivaskasi2021active} sums the log probabilities. Panoptic mapping \cite{Narita2019panopticmapping} separates meaningful objects from the background and combines the probability of the object classes by weighting based on the class confidences. All these methods propose a Bayesian update based on the probability distributions coming from the neural network. Thus, they assume that the network's predictions are correct and that the confidences are well-calibrated. In contrast, we leverage Bayesian neural networks to model the uncertainty in sensor observations and perform robust semantic fusion

\subsubsection{Bayesian Deep Learning}

Bayesian neural networks (BNNs) infer a distribution over the network weights to model the uncertainty of the training process. Since the analytical computation of the posterior is intractable, several methods propose computing approximate distributions. Monte-Carlo (MC) dropout \cite{Gal2016} and deep ensembles \cite{bayesi, lakshminarayanan2017simple} are two popular sampling techniques that approximate the true posterior using discrete samples. On the other hand, feature-space techniques \cite{lee2018simple, 
mukhoti2023deep, 
postels2020quantifying}
estimate the uncertainty with a single-pass by measuring the distance or density of the samples in feature space compared with the training data. 

BNN applications in robotic systems show multiple uses of uncertainty estimation in perception and planning for learning, but not in guiding semantic mapping. At the perception stage, the uncertainty in the model, or epistemic uncertainty, appears in regions where the model parameters are less confident or poorly estimated, leading to ambiguous or challenging pixels \cite{Gal, aff_loren} and identifies false positive detections \cite{miller2021uncertainty}. The aleatoric uncertainty, or entropy of the data distribution, stems from inherent randomness or variability in the data distribution. It is present in the contour of objects and noisy regions \cite{shridhar2018uncertainty}. Other works employ the uncertainty at the planning stage to perform active learning~\cite{lenczner2022dial}, which reduces the labeling cost by selecting the most informative samples~\cite{ruckin2022informative} and allows for domain adaptation~\cite{zurbrugg2022embodied}. Finally, epistemic uncertainty has been used in object goal navigation to make informed decisions about exploring uncertain areas or reaching known target objects \cite{georgakis2021learning}. To the best of our knowledge, few works consider epistemic uncertainty as a weight for semantic fusion in object detection tasks~\cite{feldman2018} and planning ~\cite{tchuiev2023}. In our case, we apply the epistemic uncertainty of BNNs to semantic fusion over all the pixels in the image.

\section{Method} %

Our objective is to perform robust semantic mapping using a given uncertainty-aware NN on a mobile sensor. This is accomplished by considering the model's confidence and regularizing the map fusion in order to account for re-observations and reject outliers coming from overconfident predictions. First, we describe the semantic map structure and introduce an overview of Bayesian neural networks to estimate the observations' uncertainties. Then, we introduce the novel Bayesian fusion method based on confidence weighting and regularization.

\subsection{Map Description}

We consider a generic representation for the map using a voxel set, $\mathcal{M}$, for all the detected surfaces. Each voxel, $m \in \mathcal{M}$ has a semantic label, $l_m\in \mathcal{K} = \{ 1, \dots, K \}$, where $\mathcal{K}$ is the set of pre-defined possible classes. The objective of the semantic mapping algorithm is to infer the semantic class for all the voxels.
In order to do that, the categorical probability distribution of a voxel $m$ is defined by the probability of the voxel belonging to each of the defined classes,
\begin{equation}
    p(l_m = l_i | \textbf{p}) = \prod_{j=1}^k p_j^{[j = i]},
\end{equation}
where $\mathbf{p} = (p_1, \dots , p_k)$ such that $p_j$ represents the probability of the voxel belonging to class $j$, the exponent, $[j = i],$ is the indicator function, and $\sum_{j=1}^k p_j = 1.$
Without loss of generality, we describe the whole semantic fusion process for a single voxel $m$, denoting $p(l_i) \equiv p(l_m=l_i| \mathbf{p}),$ understanding that the same procedure is applied to all the voxels concurrently.

Semantic inference is carried out aggregating all the measurements that are taken of the voxel in a Bayesian fashion. Denoting $\mathbf{X}_t$ as the measurement acquired at time $t$, the objective is to obtain the posterior distribution, $p(l_i|\mathbf{X}_{1:t})$, where the initial prior is a uniform distribution, to model the fact that there is no initial knowledge of the class. In the paper, $\mathbf{X}_t$ is the sensor input, which can be an image or a LIDAR scan, for example. Then, we let $p(\mathbf{Y}_{t}|\mathbf{X}_{t})$ be the class probability distribution associated with each data $\mathbf{Y}_{t} = f_\omega(\mathbf{X}_t)$ (pixel, LIDAR point, etc.), which is obtained from a semantic segmentation neural network $f_\omega(\cdot)$.   
Under this observation model, standard semantic Bayesian fusion~\cite{McCormac2017semanticfusion} approximates the posterior of a voxel by 
\begin{equation}
\label{Eq:ClassicFusion}
    p(l_i|\mathbf{X}_{1:t}) \propto p(l_i|\mathbf{X}_{1:t-1}) \prod_{j} p(y_j=l_i|\mathbf{X}_{t}),
\end{equation}
where the product, $\prod_j,$ includes, for the time $t,$ the NN outputs $y_j \in \mathbf{Y}_{t}$ (labeled pixels, points, etc.) from $\mathbf{X}_t$ whose projection falls in the corresponding voxel.
We assume the computation of this projection is given by existing geometric mapping algorithms, e.g., SLAM, noting that
both Eq.~\eqref{Eq:ClassicFusion} and our proposed fusion method can be applied independently of how the metric part is computed.

\begin{figure}[t]
    \centering
    \includegraphics[width=0.95\columnwidth]{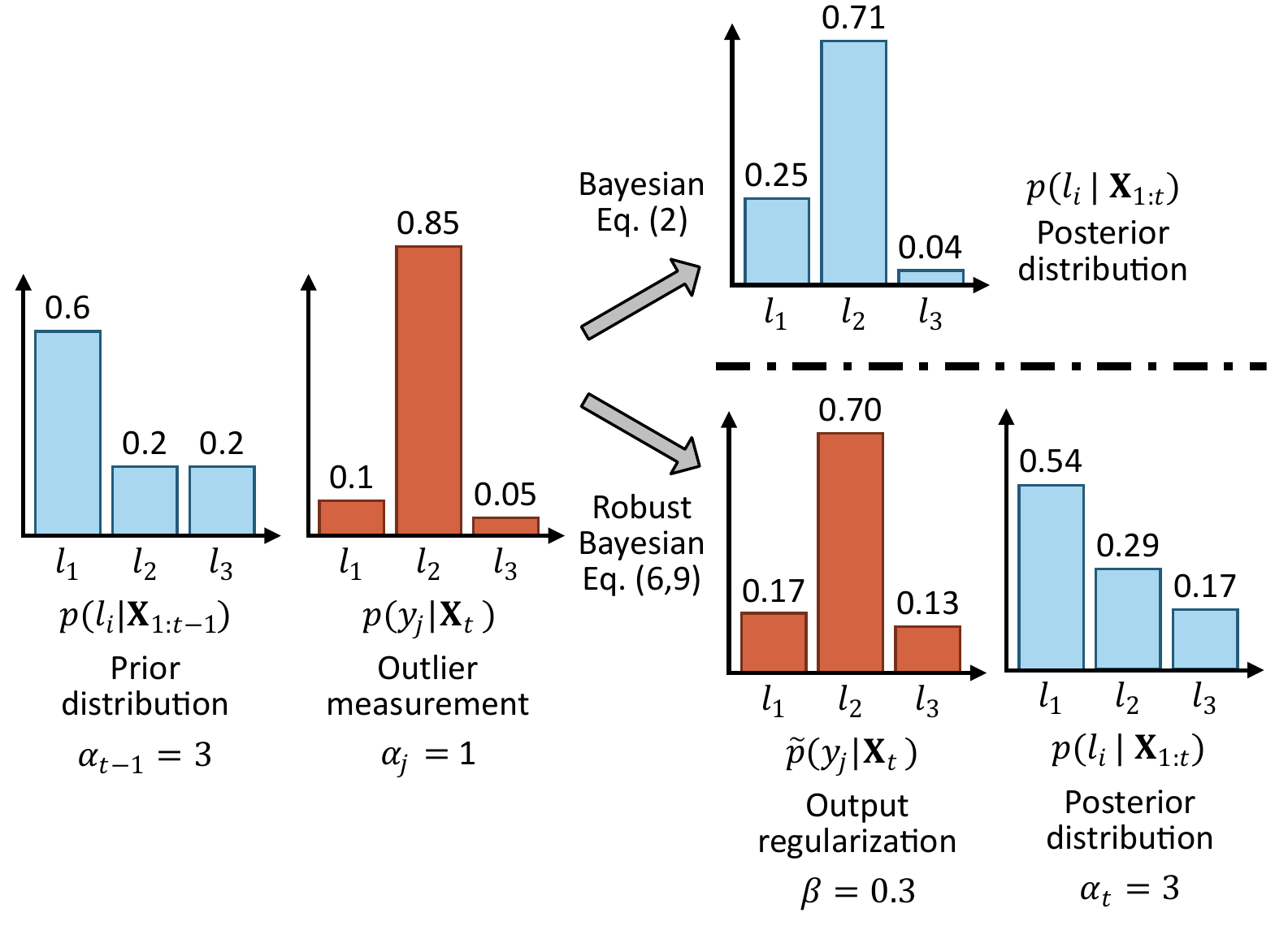}
    \caption{\footnotesize{Influence of an outlier in traditional and our robust semantic Bayesian fusion. One wrong observation (middle distribution) can shift drastically the prior distribution (left), to values where the highest probability belongs to the wrong class (right upper). Our method (right lower) first regularizes the measurements to avoid overconfidence and considers the epistemic uncertainty of the model in the Bayesian fusion with the $\alpha$ term.}}
    \label{fig:NonRobustFusion}
\end{figure}

The inclination of NNs to return over-confident observations is an important problem for~\eqref{Eq:ClassicFusion} because a single outlier with high probability has a strong negative impact in the posterior distribution.A simple numeric example of this problem is illustrated in Figure~\ref{fig:NonRobustFusion}, where it can be observed how one measurement shifts the posterior distribution to wrong values independently of the number of measurements in $\mathbf{X}_{1:t-1}$.

\subsection{BNN for Semantic Observation}

As a first step towards obtaining a more robust semantic fusion mechanism, we consider the use of a Bayesian neural network, instead of a standard neural network.

Considering a posterior distribution $p(\omega|\mathcal{D})$ on the neural network weights after training on dataset $\mathcal{D}$, the semantic observation of the network can be defined by the predictive posterior distribution:
\begin{equation}
\label{Eq:BDNNIntegral}
   p(y_j|\mathbf{X}_t,\mathcal{D}) = \int_\omega p(y_j|\mathbf{X}_t, \omega) p(\omega|\mathcal{D}) d\omega. 
\end{equation}
Without loss of generality, we use Monte-Carlo dropout to compute the posterior distribution $p(\omega|\mathcal{D})$, but other methods such as deep ensembles, or feature density could be used. We approximate the predictive posterior distribution using directly the Monte-Carlo samples
\begin{equation}
\label{Eq:BNNObservation}
    p(y_j|\mathbf{X}_t,\mathcal{D}) \approx \frac{1}{M} \sum_{i = 1}^M p(y_j|\mathbf{X}_t, \omega^{(i)}) 
\end{equation}
where $p(y_j|\mathbf{X}_t, \omega^{(i)})$ is the output of the network for sample $\omega^{(i)}$. The Monte-Carlo samples are generated at test time, sampling a Bernoulli distribution that multiplies the values of each weight in the network. In practice, this is implemented by dropout layers that remain active at test time.

Importantly for the purpose of the paper is that this definition of the observation enables the computation of the uncertainty associated with the output of the network $\Sigma(\mathbf{X}_t)$, which can be divided into $\Sigma = \Sigma_a + \Sigma_e$. In this case, $\Sigma_a(y_j)$ is called the aleatoric uncertainty, it is related to the data noise and it is already encoded in the entropy of the output probabilities. Notice that, higher $\Sigma_a(y_j)$ will have a low influence in Eq.~\ref{Eq:ClassicFusion} with the uniform distribution having no effect.
The epistemic uncertainty $\Sigma_e(y_j)$ represents the uncertainty of the model or the lack of knowledge with respect to a new input $\mathbf{X}_t$,
\begin{equation}
    \begin{split}
    \Sigma_e(y_j) = \frac{1}{M} \sum_{i=1}^{M} & \left[\Bigl(p(y_j|\mathbf{X}_t, \omega^{(i)}) - p(y_j|\mathbf{X}_t,\mathcal{D})\Bigr) \right. \\
    & \left. \Bigl(p(y_j|\mathbf{X}_t, \omega^{(i)}) - p(y_j|\mathbf{X}_t,\mathcal{D})\Bigr)^T \right].
\end{split}
\end{equation}
The lower this quantity, the more we can trust the prediction of the network about the observation.

\subsection{Robust Fusion Algorithm}

In order to reduce the sensitivity of the fusion to outliers with low aleatoric uncertainty, i.e., wrong observations with high confidence in the winning class, we first include  a constant regularization term, $\beta,$ in the output of the network,
\begin{equation}
\label{Eq:Regularization}
    \tilde{p}(y_j|\mathbf{X}_t,\mathcal{D}) \approx (1-\beta)\frac{1}{M} \sum_{i = 1}^M p(y_j|\mathbf{X}_t, \omega^{(i)}) + \beta\mathcal{U},
\end{equation}
where $\mathcal{U}= (\frac{1}{k},\ldots,\frac{1}{k})$ is a uniform distribution over the semantic class set $\mathcal{K}.$ This equation can be understood as the marginalization of the probability distribution given by the neural network, conditioned to the probability it has to make a mistake, defined by $\beta$.

The positive consequence of the regularization is that overconfident outlier measurements will have less influence on the posterior distribution. On the other hand, the mapping algorithm will require more measurements of each voxel before committing to a specific class.

Second, we modify the categorical distributions from Eq.~\eqref{Eq:ClassicFusion} by Dirichlet distributions $\mathcal{D}ir(p,\mathbf{\alpha})$, where the concentration parameters $\mathbf{\alpha}=\{\alpha_i\}_{i=1}^K$ are inversely related to the epistemic uncertainty component, that is,
\begin{equation}
\label{Eq:concentration}
    \alpha_{t,j,i} = -\log(\Sigma_e(y_j, l_i)),
\end{equation}
where $\Sigma_e(y_j, l_i))$ represents the marginal variance associated with label $l_i$.
\begin{equation}
    \bar{\alpha}_{t,i} = \max_{\tau<t,j} \{\alpha_{\tau,j,i}\}.
\end{equation}
Therefore, our Bayesian fusion equation is
\begin{equation}
\begin{split}
    p( l_i &| \mathbf{X}_{1:t}) \propto \\
    &p( l_i | \mathbf{X}_{1:t-1})^{\frac{\bar{\alpha}_{t-1,i}}{\bar{\alpha}_{t,i}}} \prod_{j} \tilde{p}(y_j=l_i|\mathbf{X}_{t},\mathcal{D})^{\frac{\alpha_{t,j,i}}{\bar{\alpha}_{t,i}}},
\end{split}
\label{Eq:FusionWeighted}
\end{equation}
where $\bar{\alpha}_{t,i}$ is a normalizing constant equal to the the maximum value up to time $t$ for label $l_i$, that is, 
\begin{equation}
    \bar{\alpha}_{t,i} = \max_{\tau<t,j} \{\alpha_{\tau,j,i}\}.
\end{equation}
We can think that the Dirichlet distribution in this case represents the confidence on the underlying distribution. For example, if we have an observation with a higher concentration, it is equivalent to fusing multiple observations with that value, therefore having a higher multiplicity in Eq.~\eqref{Eq:ClassicFusion}.
Figure~\ref{fig:NonRobustFusion} illustrates the effect of the regularization in the measurement and the effect of the fusion of Dirichlet distributions in the computation of the posterior.

\begin{figure}[t]
    \centering
    \includegraphics[width=0.98\columnwidth]{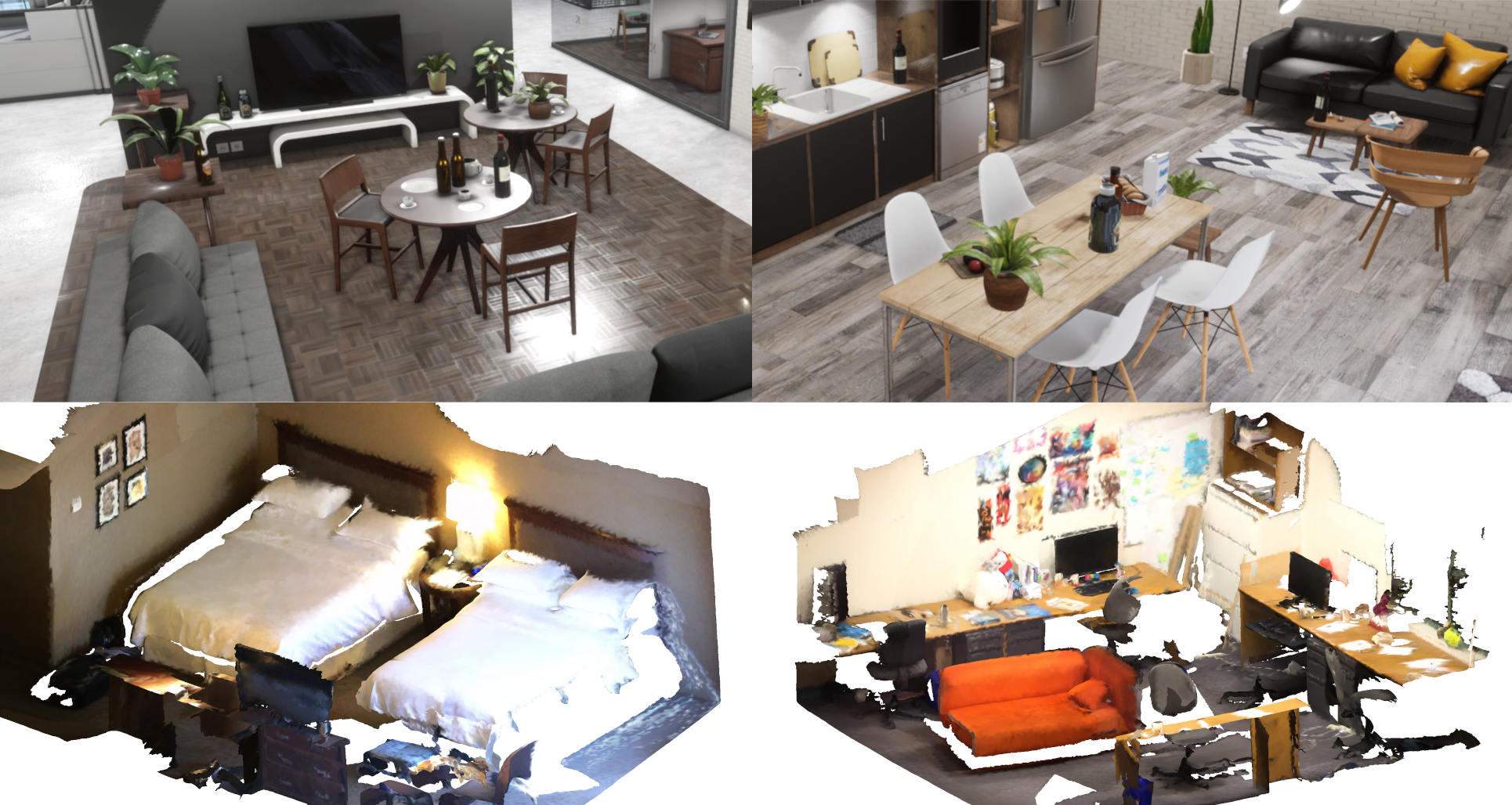}
    \caption{\footnotesize{Overview of the \textit{Office} and \textit{House} virtual scenes and two environments from the real dataset, ScanNet (sequences 6 and 40).}}
    \label{fig:environments}
\end{figure}

\section{Experiments}

We evaluate the impact of our robust fusion method on the semantic mapping task and investigate the advantages it offers compared to existing fusion methods. To achieve this, we use an RGB-D camera to capture images of different environments along a fixed trajectory with known poses, so that uncertainties and errors related to geometry do not affect the analysis. These images are then input to a Bayesian semantic segmentation network, and the resulting predictions are projected onto a map using the camera's depth information and fused considering different methods.

\subsection{Environments}

Ideally, we would like to abstract the problem of semantic fusion from other uncertainties arising from sensor pose and depth estimation. Furthermore, having accurate semantic ground truth (GT) is also advantageous.  Therefore, we initially test our semantic mapping in a photo-realistic simulation environment using Unreal Engine 4\footnote{\url{https://www.unrealengine.com/en-US}} and the Airsim simulator to model a virtual camera in the scene. The simulator provides the sensor poses, RGB images, and GT depth and semantic labels. We model two different environments called \textit{Office} and \textit{House}. Figure \ref{fig:environments} offers an overview of the environments. We conduct mapping over three different trajectories within each environment and evaluate the aggregation of the results.

We further evaluate our approach on three sequences from ScanNet (6, 9 and 40). This is a real dataset with annotated 3D reconstructions of indoor scenes that provide GT camera poses and depth measurements \cite{dai2017scannet}. This dataset validates the application of our method to real-world scenarios and how it performs under noisy depth measurements coming from a real device.

For the mapping framework, we use a simple voxel hash-map. By relaxing the need to deal with occupancy probability, we only store surface (i.e., occupied) voxels and avoid storing the \textit{free} space, reducing the storage and increasing the efficiency, resulting in a simple scheme to study the semantic fusion problem. We always use a voxel size of $0.1 \si{m}$.

\begin{figure}[t]
    \centering
    
    \includegraphics[width=0.11\textwidth]{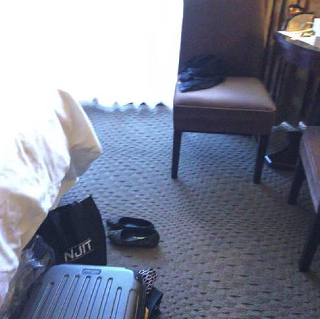}
    \includegraphics[width=0.11\textwidth]{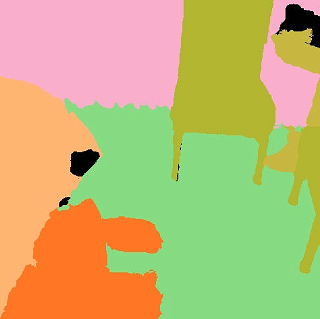}
    \includegraphics[width=0.11\textwidth]{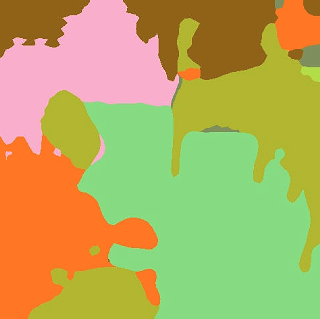}
    \includegraphics[width=0.11\textwidth]{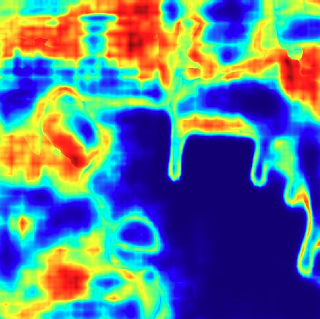} \\
    
    \includegraphics[width=0.11\textwidth]{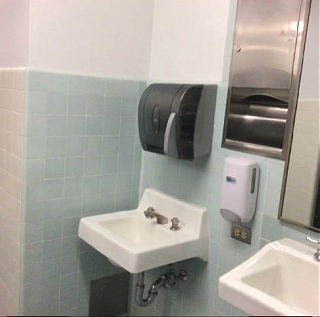}
    \includegraphics[width=0.11\textwidth]{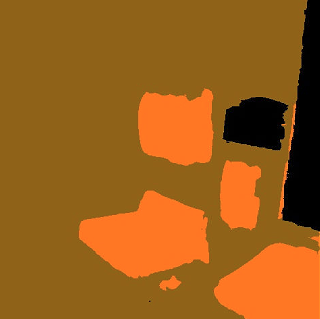}
    \includegraphics[width=0.11\textwidth]{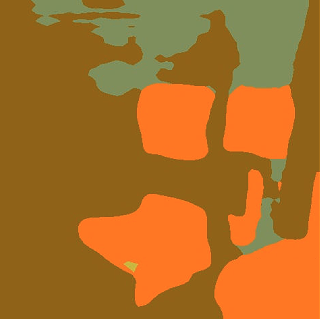}
    \includegraphics[width=0.11\textwidth]{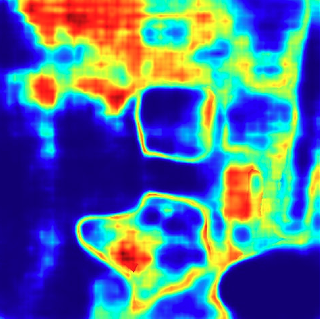} \\

    \begin{subfigure}[b]{0.11\textwidth}
         \centering
         \includegraphics[width=\textwidth]{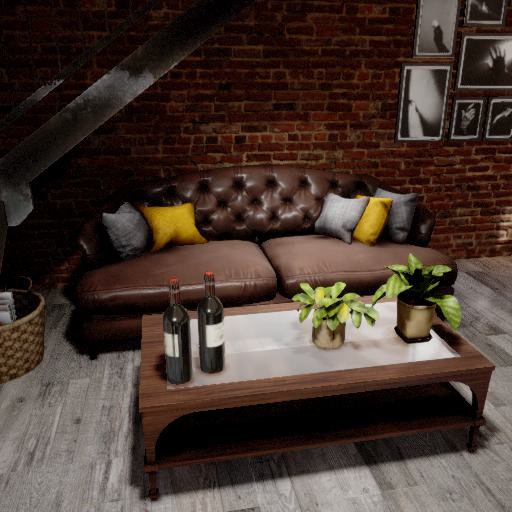}
         \caption{\begin{tabular}[c]{@{}c@{}}RGB \\ Image \end{tabular}}
         \label{fig:rgb_image}
     \end{subfigure}
     \begin{subfigure}[b]{0.11\textwidth}
         \centering
         \includegraphics[width=\textwidth]{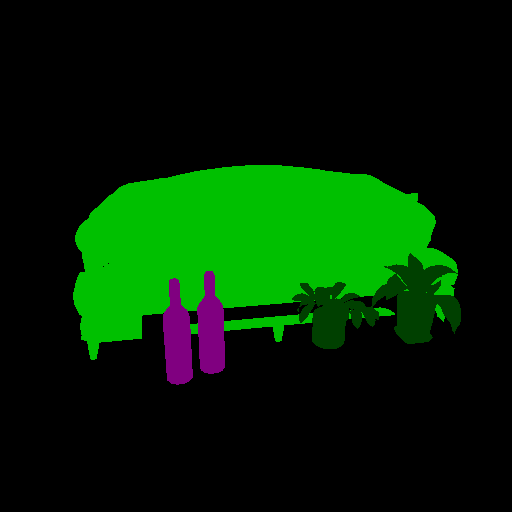}
         \caption{\begin{tabular}[c]{@{}c@{}}Ground \\ Truth \end{tabular}}
         \label{fig:ground_truth_uncerts}
     \end{subfigure}
     \begin{subfigure}[b]{0.11\textwidth}
         \centering
         \includegraphics[width=\textwidth]{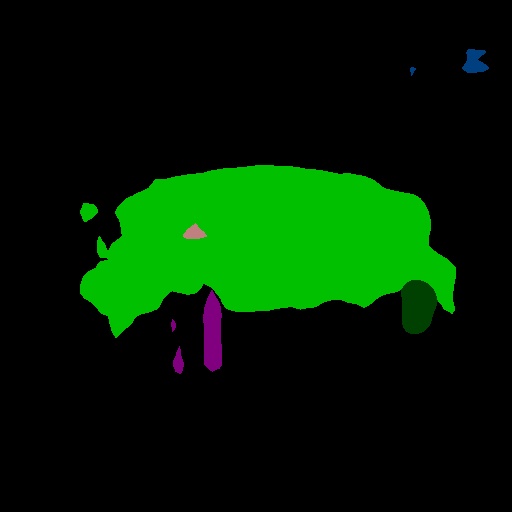}
         \caption{\begin{tabular}[c]{@{}c@{}}BNN \\ Prediction \end{tabular}}
         \label{fig:prediction_uncerts}
     \end{subfigure}
     \begin{subfigure}[b]{0.11\textwidth}
         \centering
         \includegraphics[width=\textwidth]{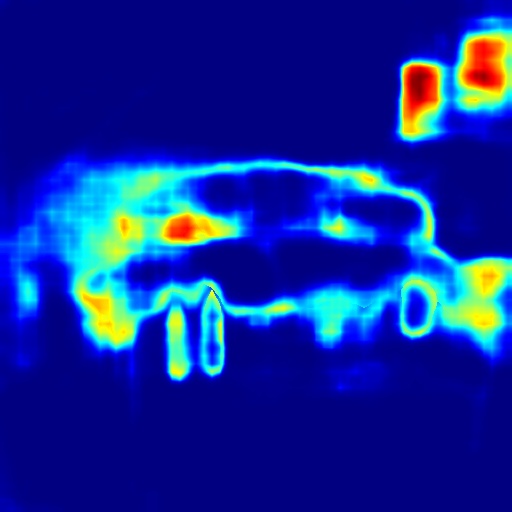}
         \caption{\begin{tabular}[c]{@{}c@{}}Epistemic \\ Uncertainty \end{tabular}}
         \label{fig:epist_uncerts}
     \end{subfigure} \\

    \caption{\footnotesize{Estimation of the BNN in scenes of the ScanNet (top) and our simulated environments (down). We find higher values of the epistemic uncertainty (in red) in the regions where the BNN fails in the prediction, showing its degree of confidence.}}
    \label{fig:BNN_outs}
\end{figure}
\begin{table*}[t]
\caption{\footnotesize{Quantitative results on the virtual environment. We report the IoU per class, the mIoU, and the mAcc to evaluate the quality of the semantic mapping.  We aggregate the measurements from three trajectories in each of the two environments.}}
\label{Tab:ResultsVirtual}
\centering
\begin{tabular}{cc|ccccccc|cc}
\multicolumn{2}{c|}{Method} & \cellcolor[HTML]{343434}{\color[HTML]{EFEFEF} Back} & \cellcolor[HTML]{604BA4}{\color[HTML]{EFEFEF} Bottle} & \cellcolor[HTML]{FE0000}{\color[HTML]{EFEFEF} Chair} & \cellcolor[HTML]{F56B00}{\color[HTML]{EFEFEF} Table} & \cellcolor[HTML]{009901}{\color[HTML]{EFEFEF} Plant} & \cellcolor[HTML]{34FF34}{\color[HTML]{000000} Sofa} & \cellcolor[HTML]{00009B}{\color[HTML]{EFEFEF} TV} & {\color[HTML]{000000} mIoU} & {\color[HTML]{000000} Acc.} \\ \hline
\multicolumn{2}{c|}{Sum. Probs. \cite{McCormac2018fusionplusplus}} & 89.2 & 25.4 & 30.7 & 12.5 & 18.0 & 43.7 & 6.4 & 32.3 & 89.1 \\ \hline
\multicolumn{2}{c|}{Sum. Labels. \cite{Schmid2022multitsdf}} & 89.9 & 27.5 & 32.3 & 15.1 & 19.9 & 47.6 & 7.2 & 34.2 & 89.5 \\ \hline
\multicolumn{2}{c|}{Bayesian \cite{McCormac2017semanticfusion}} & 90.5 & 20.6 & 44.6 & 21.9 & 18.1 & 51.5 & 5.3 & 36.1 & 90.8 \\ \hline
 & R & 91.1 & 26.9 & \textbf{46.5} & \textbf{23.2} & 20.1 & 58.2 & 9.9 & 39.4 & 91.4 \\
 & D & 91.4 & 24.1 & 42.7 & 21.3 & 19.8 & 68.9 & 3.4 & 38.8 & 91.8 \\
\multirow{-3}{*}{\begin{tabular}[c]{@{}c@{}}Robust Fusion\\ (Ours)\end{tabular}} & D + R & \textbf{91.9} & \textbf{30.7} & 45.5 & 22.4 & \textbf{21.5} & \textbf{71.2} & \textbf{10.6} & \textbf{42.0} & \textbf{92.2}
\end{tabular}
\end{table*}
\begin{table*}[t]
\caption{\footnotesize{Quantitative results on the ScanNet dataset. We report the IoU per class, the mIoU and the mAcc to evaluate the quality of the semantic mapping.  We aggregate the measurements from three different scenes.}}
\centering
\resizebox{\textwidth}{!}{%
\begin{tabular}{cc|cccccccccccc|cc}
Method & &
  \cellcolor[HTML]{FFBB78}Bed &
  \cellcolor[HTML]{4E47B7}\color[HTML]{EFEFEF} Ceiling &
  \cellcolor[HTML]{BCBD22}Chair &
  \cellcolor[HTML]{98DF8A}Floor &
  \cellcolor[HTML]{8C9965}\color[HTML]{EFEFEF} Furniture &
  \cellcolor[HTML]{FF7F0E}Objects &
  \cellcolor[HTML]{A1AB1B}Picture &
  \cellcolor[HTML]{BEE140}Sofa &
  \cellcolor[HTML]{CEBE3B}Table &
  \cellcolor[HTML]{73B0C3}TV &
  \cellcolor[HTML]{996C06}\color[HTML]{EFEFEF} Door &
  \cellcolor[HTML]{F7B6D2}Window &
  mIoU &
  Acc. \\ \hline
Sum. Probs \cite{McCormac2018fusionplusplus}     &   & 22.9 & 16.8 & 21.4 & 48.6 & 32.7 & 20.4 & 5.2 & 24.2 & 15.4 & 14.1 & 59.2 & 10.2 & 20.8 & 53.1\\ \hline
Sum. Labels \cite{Schmid2022multitsdf}       &        & 27.5 & 19.6 & 22.3 & 51.1 & 32.0 & 20.2 & 5.8 & 23.6 & 15.3 & \textbf{33.2} & 59.7 & 11.2 & 22.9 & 53.8\\ \hline
Bayesian \cite{McCormac2017semanticfusion}   &        & 29.3 & 21.7 & 26.9 & 59.7 & 35.4 & 23.5 & 7.4 & 33.5 & 15.7 & 31.9 & 63.6 & 11.7 & 25.8 & 57.9 \\ \hline
  &    R                                                & 30.9 & 22.0 & 27.5 & 60.1 & 35.9 & 23.6 & 7.4 & 34.7 & 15.8 & 28.9 & 63.6 & 12.4 & 25.9 & 58.0 \\
 &    D                                                 & 41.6 & \textbf{28.2} & \textbf{33.3} & 72.4 & 37.5 & 28.6 &\textbf{ 23.9} & \textbf{44.5} & 18.5 & 14.1 & 65.7 & \textbf{32.2} & \textbf{30.1} & 63.7 \\
\multirow{-3}{*}{\begin{tabular}[c]{@{}c@{}}Robust \\ Fusion\\ (Ours)\end{tabular}} &    R+D     &   \textbf{42.1}  &  \textbf{28.2} & 32.8 & \textbf{72.5} & \textbf{38.2} & \textbf{28.7} & \textbf{23.9} & 43.8 & \textbf{19.0} & 12.6 & \textbf{66.0} & 31.4 & 30.0 & \textbf{63.9} \\ 
\label{tab:real_to_real_exp}
\end{tabular}
}
\end{table*}
\begin{figure*}[t]
    \vspace{-10pt}
    \centering
    \begin{subfigure}[b]{0.27\textwidth}
         \centering
         \includegraphics[width=\textwidth,trim={1cm 5cm 0 2cm},clip]{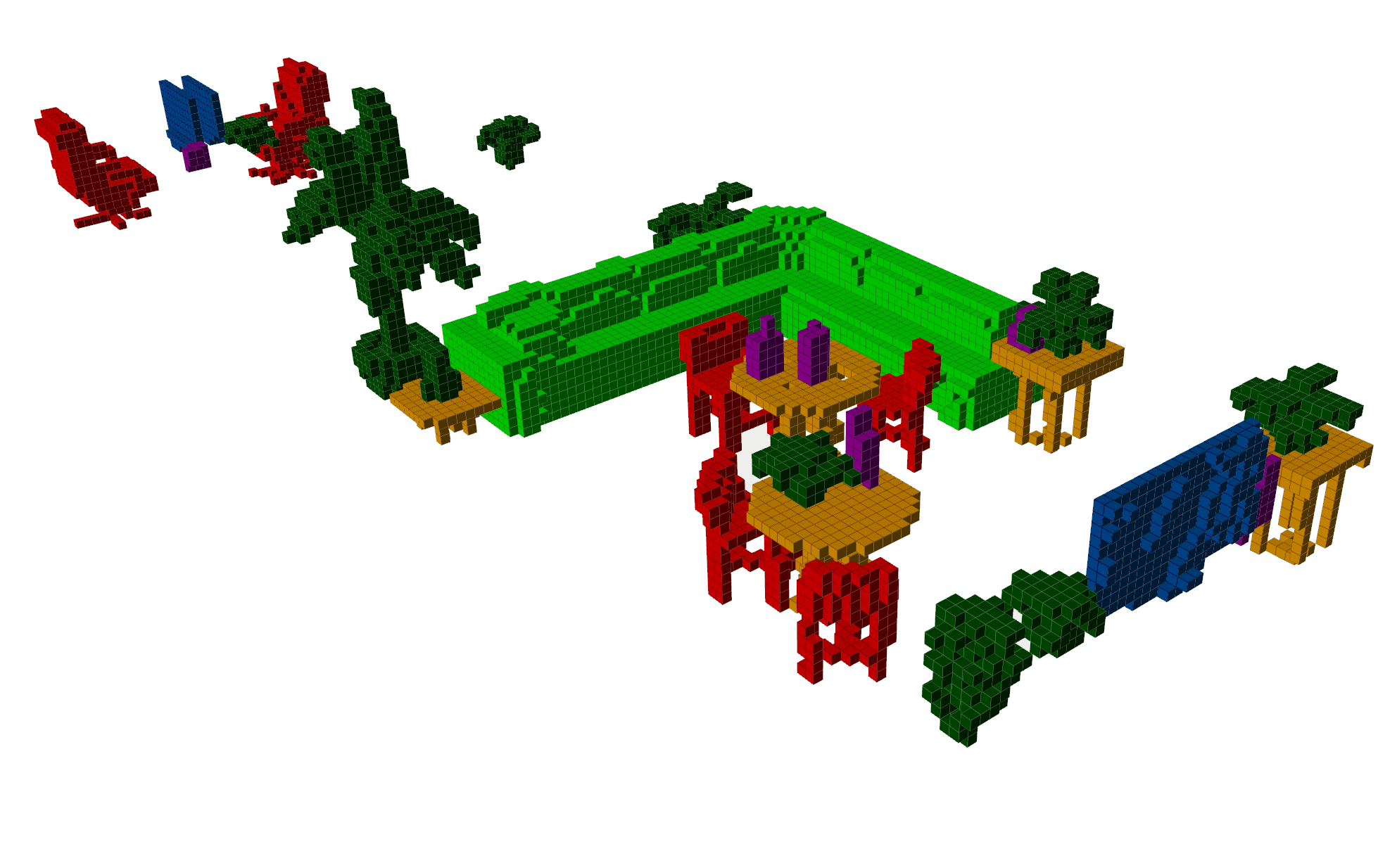}
         \label{fig:gt_virtual}
     \end{subfigure}
     \begin{subfigure}[b]{0.27\textwidth}
         \centering
         \includegraphics[width=\textwidth,trim={0 5cm 0 2cm},clip]{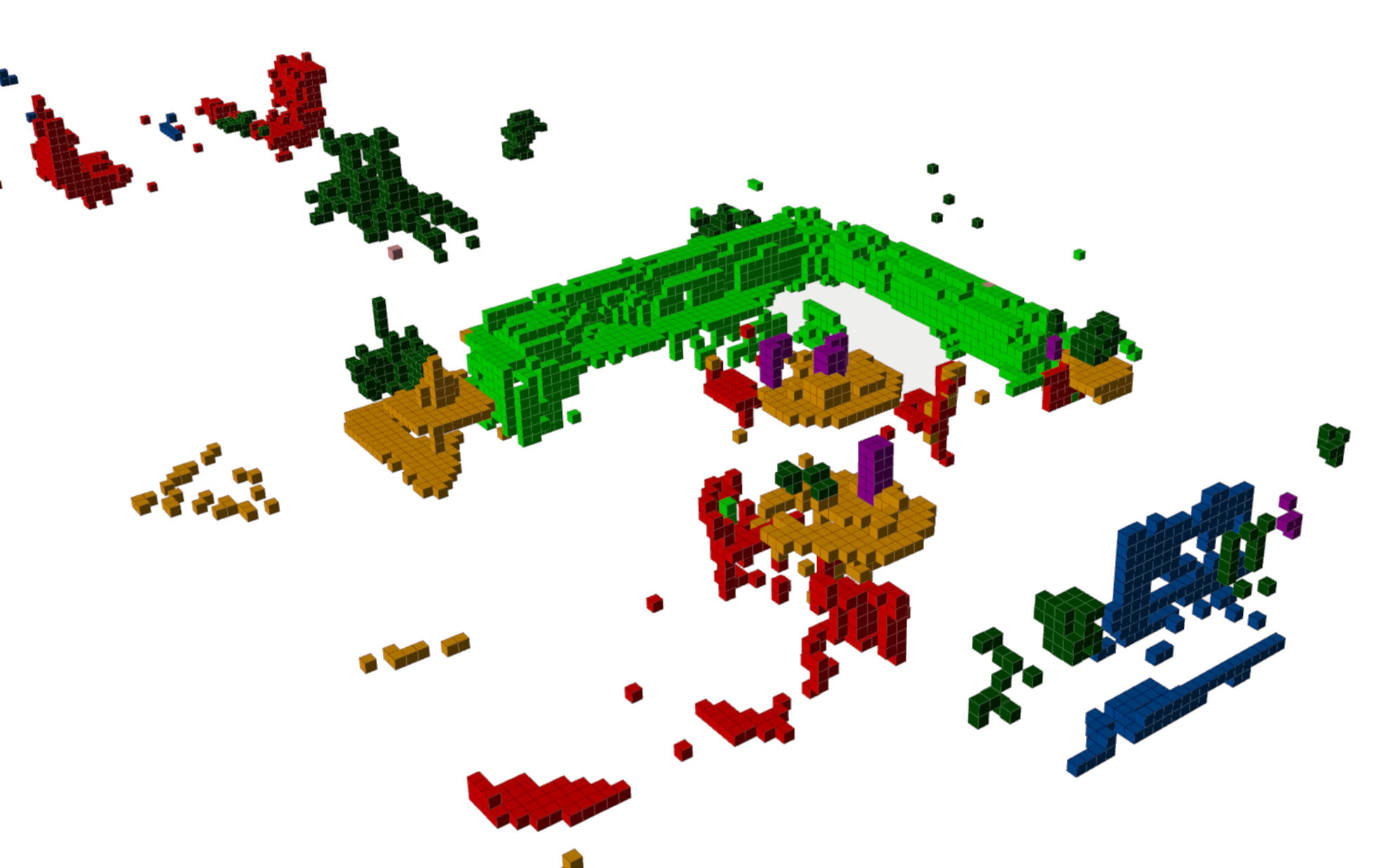}
         \label{fig:det_virtual}
     \end{subfigure}
     \begin{subfigure}[b]{0.27\textwidth}
         \centering
         \includegraphics[width=\textwidth,trim={0 5cm 0 2cm},clip]{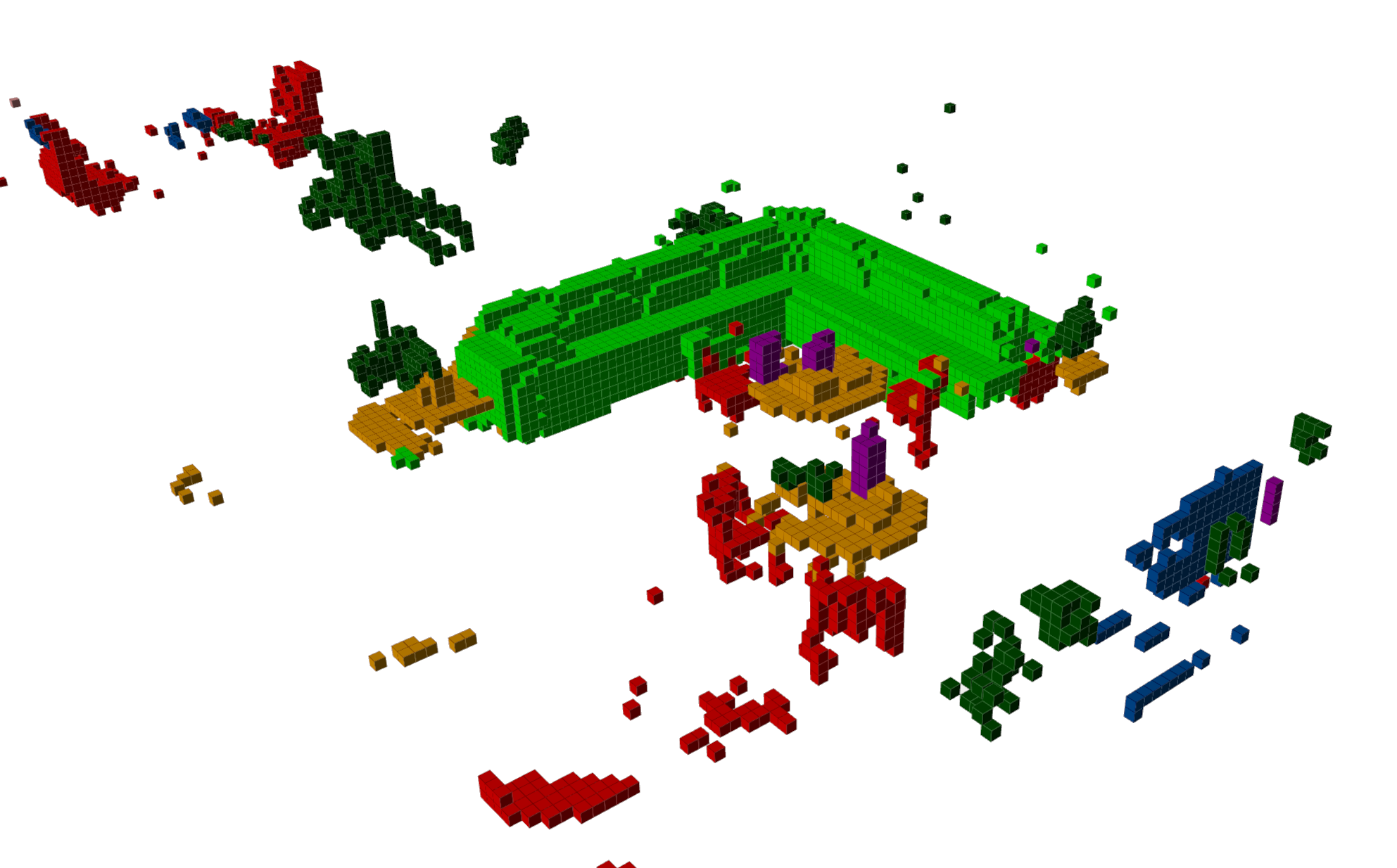}
         \label{fig:bay_virtual}
     \end{subfigure} \\

    \begin{subfigure}[b]{0.27\textwidth}
         \centering
         \includegraphics[width=\textwidth,trim={0 1cm 0 0},clip]{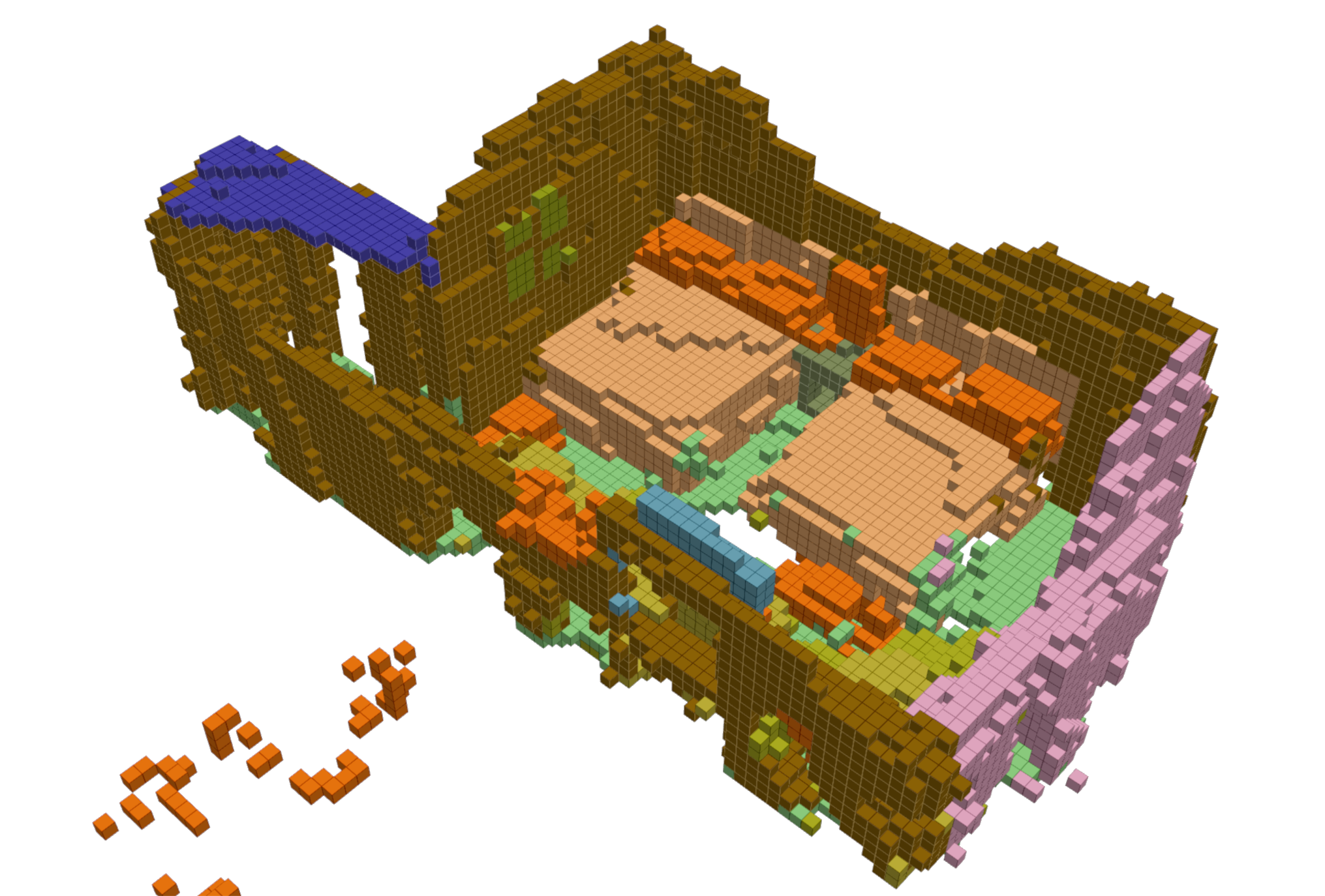}
         \caption{Ground-truth}
         \label{fig:gt_real}
     \end{subfigure}
     \begin{subfigure}[b]{0.27\textwidth}
         \centering
         \includegraphics[width=\textwidth,trim={1cm 1cm 0 0},clip]{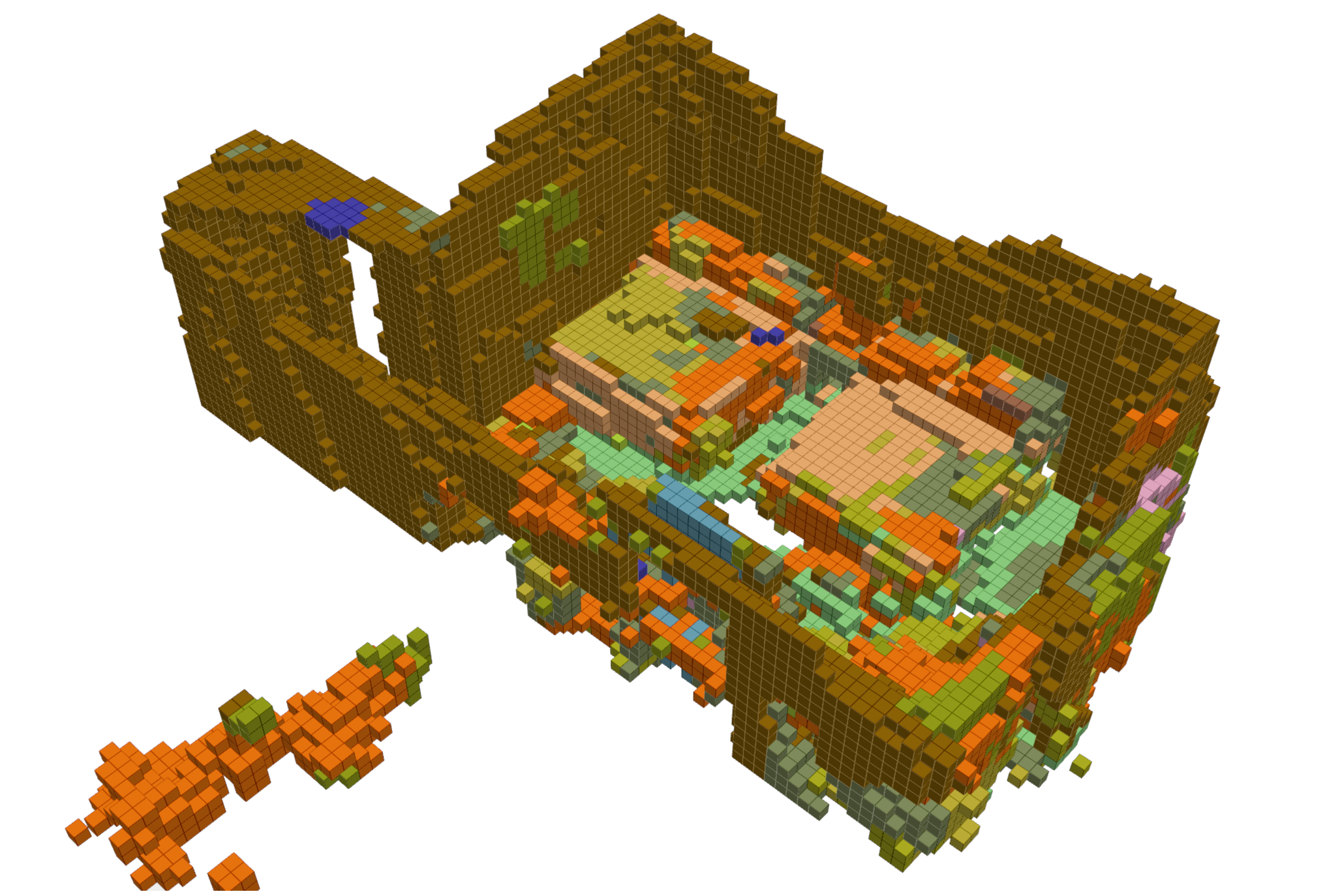}
         \caption{Standard fusion}
         \label{fig:det_real}
     \end{subfigure}
     \begin{subfigure}[b]{0.27\textwidth}
         \centering
         \includegraphics[width=\textwidth,trim={0 1cm 1cm 0},clip]{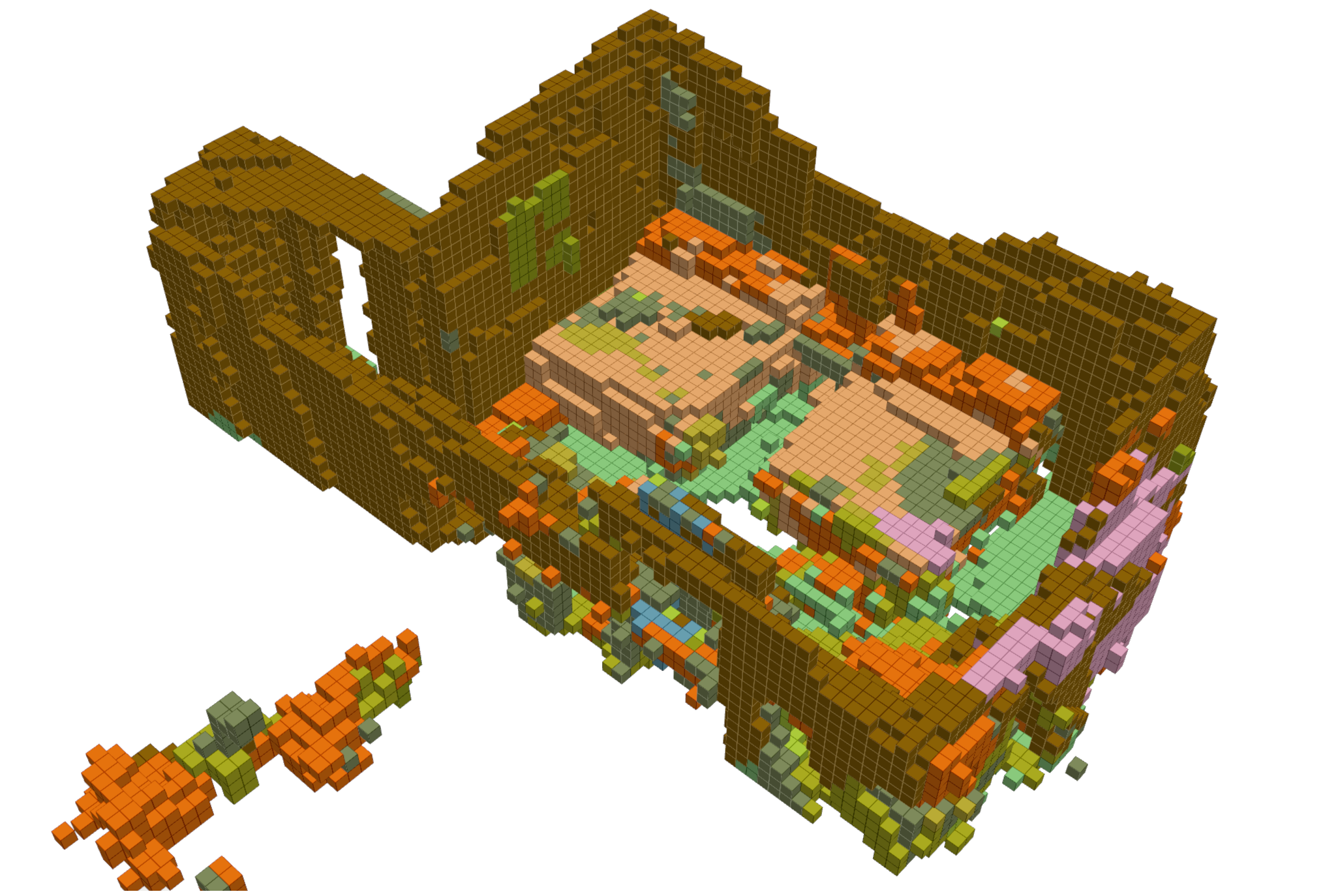}
         \caption{Robust fusion (ours)}
         \label{fig:bay_real}
     \end{subfigure} \\

    \caption{\footnotesize{Qualitative examples of the experiments performed in virtual (top) and real (bottom) scenarios. Voxels corresponding to the \textit{background} class were removed for clarity. Details can be observed in the supplementary material. Our robust Bayesian fusion method is able to improve the mapping by reducing the influence of wrong measurements on the map.}}
    \label{fig:results_quali}
    \vspace{-15pt}
\end{figure*}
\subsection{Sensor configuration}

We use a BNN based on the semantic segmentation model DeepLab-v3 \cite{chen2017rethinking}. We place a dropout layer after each ResNet block with a dropout rate $d=0.3,$ following \cite{aff_loren}. At inference, we keep the dropout layers active and obtain $32$ samples from the approximate posterior using MC dropout. 
For the baseline models, we use the same network with inactive dropout layers, as in traditional deterministic models.

Real applications require robust methods that can generalize to new situations. We introduce this challenge in our experiments by training on different environments than that of the testing. For the simulated environment, we train on real data using the VOC12 Dataset~\cite{pascal-voc-2012}. We select the samples that contain any of the six classes of interest (see Table~\ref{Tab:ResultsVirtual}), corresponding to indoor objects present in the simulation setup. For the real environment, we train on the NYU-v2 dataset~\cite{Silberman:ECCV12} with 12 labels and evaluate the semantic mapping in sequences from the ScanNet dataset~\cite{dai2017scannet}.

\subsection{Metrics and Baselines}

In order to assess the quality of the mapping, we measure the Intersection over Union (IoU) over each of the classes and compute the mean IoU (mIoU). We also compute the Accuracy for all the voxels in the map. To evaluate each part of the robust fusion, we perform an ablation study using the regularization (R), the Dirichlet model (D), and both at the same time (D+R). The parameter $\beta$ used in the regularization is set to $0.3$ for all the experiments. Adapting this value depending on the environment and the class might improve the results and will be explored in the future. We also compare our method with three current semantic fusion methods using a deterministic version of our neural network: summing the predicted probabilities \cite{McCormac2018fusionplusplus}, counting the predicted classes as labels \cite{Schmid2022multitsdf}, and the Bayesian fusion \cite{McCormac2017semanticfusion} described in~\eqref{Eq:ClassicFusion}.

\subsection{Results}

Figure~\ref{fig:BNN_outs} (a) shows different examples of images in the real and simulated sequences. There, it is possible to see the difference between the GT semantic labels (b) and the output of the BNN (c), which motivates the need for better fusion models. Finally, we show the epistemic uncertainty associated with the predictions in (d). Notably, the red areas in (d), which correspond to pixels with high uncertainty, are associated in many cases with errors in the classification.

\paragraph{Simulated environments}

In Table~\ref{Tab:ResultsVirtual}, we present the quantitative results for the virtual environment. All baselines achieve comparable mIoU and accuracy, which is expected since the neural network accuracy will be similar. However, the fusion approach has a strong influence on the final map. The ablation rows show that, when applied individually, each term in our fusion (D and R) improves the quality of the semantic fusion, making them a valuable addition to the current state-of-the-art methods. Finally, the combination of D and R obtains the best results for the majority of classes.

\paragraph{Real environments}

The quantitative results on the real scenario of the ScanNet dataset are shown in Table \ref{tab:real_to_real_exp}. These results show a similar trend as those obtained for the simulated environment, showing that our approach generalizes well to real scenarios. Notably, for the case of these scenes, the contribution of the epistemic term (D) is more significant than in the simulated case. We argue that this might be produced by a larger fraction of out-of-distribution samples than in the simulated environments.

Finally, we show qualitative results of the generated maps in Figure \ref{fig:results_quali} for both configurations. Compared to the baseline, our mapping is less sensitive to outliers. For example, the sofa on the top example and the beds on the bottom are better segmented with less spurious classes. Overall, our approach shows promise for practical applications in real-world scenarios, demonstrating its potential to enhance the performance of semantic mapping in new environments, dealing with unknown data and outliers in a reliable manner.

\section{Conclusion}

In this work, we proposed a robust fusion method for semantic mapping that leverages Bayesian neural networks to consider the uncertainty of the network in the semantic fusion process. We achieved this by combining a regularization term, to mitigate the overconfidence in the predictions, together with a Dirichlet representation of the observations, using the epistemic uncertainty as a concentration parameter. Our method showed advantages over currently used methods when evaluated in both virtual photo-realistic and real environments, suggesting the importance of considering model uncertainties in tasks that require semantic understanding of the scene. Our next steps will focus on exploiting our Bayesian semantic mapping framework to actively move the sensor toward informative points of view that yield better confidence from the network, improving the accuracy and robustness of semantic mapping in real-world scenarios.

\balance

\bibliographystyle{./IEEEtran} %
\bibliography{tex/references}
\end{document}